\documentclass{article}
\usepackage{spconf,amsmath,graphicx}

\usepackage{amsfonts}
\usepackage{booktabs}

\title{Self-Supervised Path Planning in UAV-aided Wireless Networks based on Active Inference}
\name{Ali Krayani\textsuperscript{1,2}, Khalid Khan\textsuperscript{1}, Lucio Marcenaro\textsuperscript{1,2}, Mario Marchese\textsuperscript{1,2} and Carlo Regazzoni\textsuperscript{1,2}
}
\address{\small \textsuperscript{1}University of Genoa, Italy; 
\ \textsuperscript{2}Italian National Inter-University Consortium for Telecommunications (CNIT), Italy
}

\begin{document}
\ninept
\maketitle
\begin{abstract}
This paper presents a novel self-supervised path-planning method for UAV-aided networks. First, we employed an optimizer to solve training examples offline and then used the resulting solutions as demonstrations from which the UAV can learn the world model to understand the environment and implicitly discover the optimizer's policy. UAV equipped with the world model can make real-time autonomous decisions and engage in online planning using active inference. During planning, UAV can score different policies based on the expected surprise, allowing it to choose among alternative futures. Additionally, UAV can anticipate the outcomes of its actions using the world model and assess the expected surprise in a self-supervised manner. Our method enables quicker adaptation to new situations and better performance than traditional RL, leading to broader generalizability.
\end{abstract}
\section{Introduction}
\label{sec:intro}
Unmanned aerial vehicles (UAVs) have exceptional manoeuvrability, a high likelihood of establishing air-to-ground connections, and improved transmission link performance \cite{8579209, 8501974}. They can act as flying base stations and are easily relocated, making them highly beneficial in commercial, civilian, and natural disasters \cite{8531711, 10118877}. Optimizing UAV trajectories is crucial to fully harness their potential in developing future wireless systems \cite{9234110}.
Traditional methods rely on precise information about the system to design a successful UAV trajectory, which might not always be practical in real-world situations \cite{10086052}. AI techniques, such as machine learning (ML) and reinforcement learning (RL), can address challenges related to sequential decision-making, equipping UAVs with remarkable self-awareness and transforming wireless communications \cite{9741304}. However, most ML and RL methods cannot adjust to new situations, requiring extensive retraining efforts, which pose challenges for real-time prediction and decision-making \cite{9120668}. 
%
Fortunately, active inference offers a powerful alternative methodology and mathematical framework for comprehending how living organisms interact with their surroundings \cite{10.1162/NECO_a_00912}. It models perception, learning, and decision-making, aiming to maximize Bayesian model evidence or minimize free energy \cite{friston_2013}. By assessing multiple hypotheses, agents can achieve the desired outcomes.
%

Motivated by the above discussion and previous work \cite{s23156873}, we propose a method for self-supervised path planning for UAV-aided wireless networks. Our method employs the concept of active inference and comprises two main stages. Firstly, we learn a world model from demonstrations provided by an offline path planning optimizer. This enables the UAV to comprehensively understand the environment and deduce the optimizer's strategy for solving a particular task in a self-supervised manner. Secondly, we use the learned world model as an internal generative model enriched with active states to simulate the environment and plan actions that minimize the agent's surprise during online decision-making. This approach enables the UAV to navigate its surroundings with a reference model representing the goal, choosing actions that minimize unexpected or unusual observations (surprise) measured by how much they deviate from the expected goal.
The main contributions of this paper are as follows:
It expands on previous research \cite{s23156873} by exploring online planning, a prospective form of cognition. Decision-making using online planning involves taking into account the future observations that are expected to be gathered and providing guidance on how to act accordingly. We introduce the concept of "expected surprise" as a means of scoring different policies for planning. The agent performing online planning, equipped with the world model, can anticipate the outcomes of its actions, including assessing the expected surprise.
The proposed method's efficacy was evaluated across various testing scenarios with time-varying configurations. Our method surpassed the modified Q-learning approach, offering faster, more stable, and reliable solutions while demonstrating exceptional generalization proficiency.

\section{System Model and Problem Formulation}
\label{sec:systemModel}
As illustrated in Fig.1, we consider a UAV-aided wireless network composed of one UAV and $N$ hotspot areas randomly distributed on the ground. Each hotspot area has $K_{n}$ ($n \in N$) ground users (GUs) requesting data service. The UAV aims to find the best route from an initial location to visit each hotspot, provide data service to hotspot users, and then return to the starting location within a specific time period $T$ which is divided into $M$ time slots with $t$ duration each, i.e., $T=Mt$. The UAV path at time slot $t$, while flying at altitude $h_{u}$ and constant velocity $v$, can be denoted as $\mathbf{q}_{u}(t)=[x_{u}(t), y_{u}(t), h_{u}]$ and must satisfy: $\mathbf{q}_{u}(1)=\mathbf{q}_{u}(M)$ and $\vert\vert \mathbf{q}_{u}(t) - \mathbf{q}_{u}(t-1) \vert\vert^{2}\le (vt)^{2}$. 
Furthermore, the flight duration is segmented into a series of $E$ events, where each event is triggered upon the UAV's arrival at a new hotspot. The designated sequence of hotspots targeted during the flight mission is identified as $\mathbf{p}_{u}(e)=[C_{u}(e)]$ where $C_{u} \in \{1,\dots,N\}$. The probability of moving towards the next hotspot $C_{u}(e+1)$ from the current hotspot $C_{u}(e)$ can be represented by $\mathrm{Pr}(C_{u}(e+1)|C_{u}(e), \tau_{C_{u}(e+1)})$. Here, $C_{u}(e)$ is visited at time $T-\tau_{C_{u}(e)}$ and $\tau_{C_{u}(e+1)}$ indicates the remaining time to return to the initial location after serving $C_{u}(e+1)$.
Resource blocks (RB) are allocated to GUs within a specific hotspot through orthogonal frequency division multiple access (OFDMA). The achievable sum-rate in each hotspot can be then calculated as follows:
\begin{equation}
    R_{n} = \sum_{k=1}^{K_{n}} B_{k} \log_{2}\big(1+ \frac{p_{k}g_{k,u}(t)}{\sigma^{2}}\big),
\end{equation}
where $B_{k}$, $p_{k}$ represent the RB's bandwidth and transmitted power of a GU, respectively, in hotspot $n$. $\sigma^{2}=B_{k}N_{0}$ is the AWGN power spectral density, and $g_{k,u}(t)$ represents the probabilistic channel gain between UAV and GU calculated as \cite{8053918}:
$ g_{k,u}(t) = \frac{1}{\mathbb{K}_{0} d_{k,u}^{\alpha}(t)} [\mathrm{Pr}_{\mathrm{LoS}} \mu_{\mathrm{LoS}} + \mathrm{Pr}_{\mathrm{NLoS}} \mu_{\mathrm{NLoS}}]^{-1}$.
%
%
The value of $\mathbb{K}_{0}$ is determined by $\mathbb{K}_{0}=\big(\frac{4 \pi f_{c}}{c}\big)^{2}$, where $f_{c}$ is the frequency of the carrier wave and $c$ is the speed of light. $d_{k,u}$ is the 3D distance between GU and UAV and $\alpha$ is the path loss exponent. The probabilities of LoS and NLoS are represented by $\mathrm{Pr}_{\mathrm{LoS}}$ and $\mathrm{Pr}_{\mathrm{NLoS}}$, respectively. The variables $\mu_{\mathrm{LoS}}$ and $\mu_{\mathrm{NLoS}}$ represent extra factors of attenuation for LoS and NLoS transmissions, respectively, beyond the free-space propagation.

The problem statement indicates that the UAV has two primary objectives: maximizing the sum-rate and minimizing travel distance while moving between active hotspots, all while completing the task in the shortest possible time. This transforms the problem into a 'travelling salesman problem with profits' (TSPWP) \cite{doi:10.1287/trsc.1030.0079}. The TSPWP aims to find the best sequence of hotspots to visit to maximize net profit. The latter is determined by subtracting the total tour cost from the total profit earned from visiting hotspots and the total tour cost can be calculated as the total Euclidean distance covered during the tour. The wireless network can be represented as a graph $\mathcal{G}=(\mathcal{V},\mathcal{E})$, with hotspots as nodes and edges as local paths between them.
The set of vertices in the graph is denoted by $\mathcal{V} = \{v_1, \dots, v_N\}$ and the set of edges is denoted by $\mathcal{E}$. The center of $v_n$ is represented by $\mathrm{\boldsymbol{p}}_n=[\mathrm{x}_{n}, \mathrm{y}_{n}]$, and $R_{n}$ denotes the profit associated with $v_n$. Additionally, there is a cost $c_{ij}=d(\mathrm{\boldsymbol{p}}_{i}, \mathrm{\boldsymbol{p}}_{j}) = \sqrt{(\mathrm{x}_{i} - \mathrm{x}_{j})^{2} + (\mathrm{y}_{i} - \mathrm{y}_{j})^{2}}$ associated with each edge $(v_i, v_j) \in \mathcal{E}$. Therefore, the objective function of the problem can be expressed as:
\begin{subequations}
    \begin{align}
        \min \quad & \alpha \sum_{(v_{i},v_{j}) \in \mathcal{E}} c_{ij}x_{ij} - \beta \sum_{v_{j} \in \mathcal{V}}^{} R_{j} y_{j}, \label{TSPWP_objFunc_constraint1} \\
        \textrm{s.t.} \quad & \sum_{\substack{v_{i} \in \mathcal{V} \\ v_{j} \in \mathcal{V} \setminus \{v_{i}\}}} x_{ij} = y_{i}, \label{TSPWP_objFunc_constraint2} \\
        & \sum_{\substack{ v_{j} \in \mathcal{V} \\ v_{i} \in \mathcal{V} \setminus \{v_{j}\}}} x_{ij} = y_{j}, \label{TSPWP_objFunc_constraint3} \\
        & x_{ij} \in \{0,1\}, \ (v_{i},v_{j}) \in \mathcal{E}, \label{TSPWP_objFunc_constraint4} \\
        & y_{ij} \in \{0,1\}, \ (v_{i} \in \mathcal{V}), \label{TSPWP_objFunc_constraint5} \\
        & \alpha + \beta = 1.
    \end{align}
\end{subequations}
\begin{figure}[t!]
    \centering
    \includegraphics[width=5.3cm]{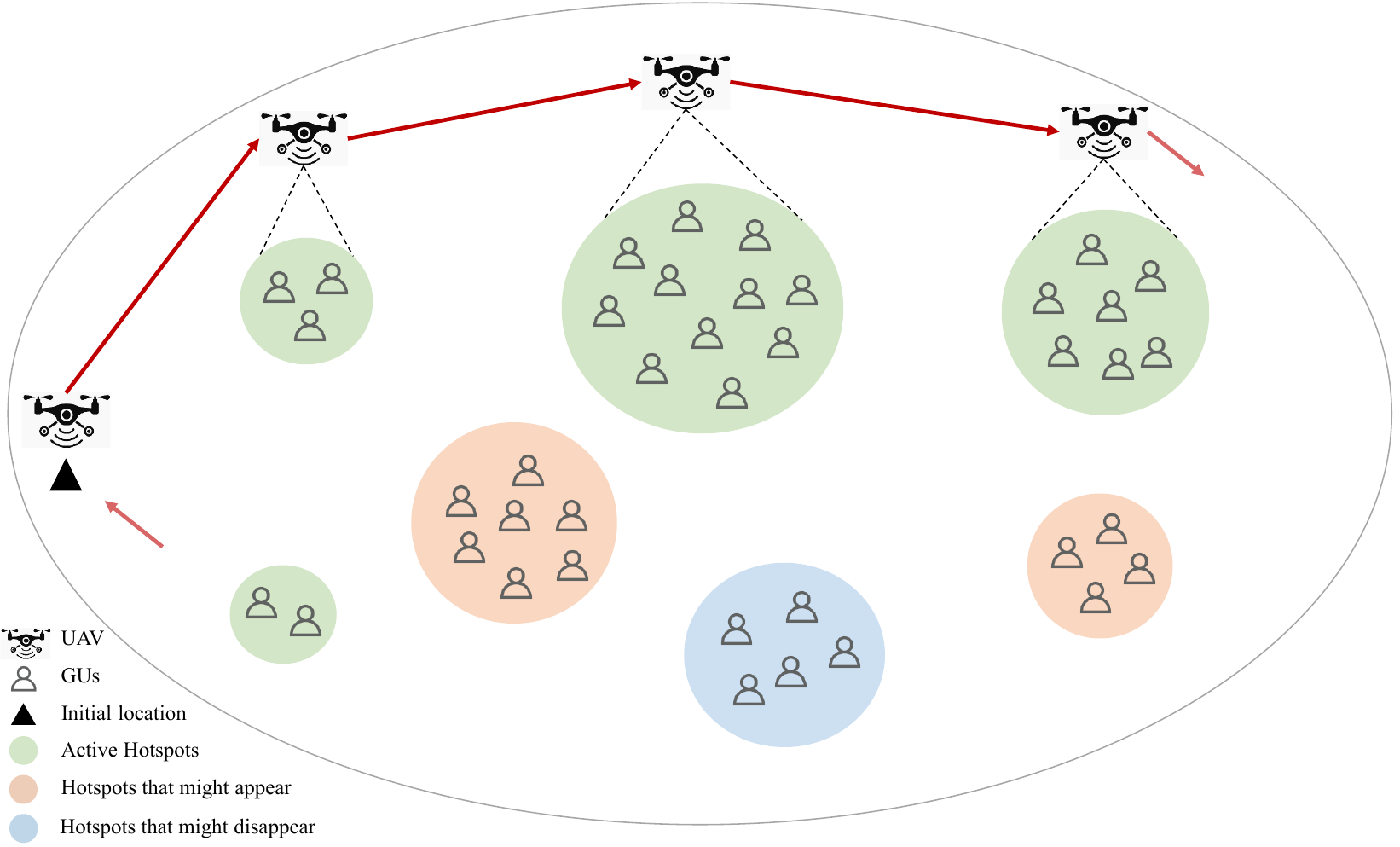}
    \caption{System Model.}
    \label{fig:systemModel}
\end{figure}
The constraints \eqref{TSPWP_objFunc_constraint2} and \eqref{TSPWP_objFunc_constraint3} refer to the assignment of edges and vertices in the solution. Binary variable $x_{ij}$ is associated with edge ($v_{i}$, $v_{j}$) and is set to $1$ only if ($v_{i}$, $v_{j}$) is used in the solution. Binary variable $y_{i}$ is associated with vertex $v_{i} \in V$ and is set to $1$ only if $v_{i}$ is visited.

It is important to note that our paper's objective is not to directly solve the TSPWP problem (i.e., maximizing sum-rate and minimizing completion time). Instead, we aim to employ an optimizer that can directly solve the problem and use the resulting solutions as demonstrations for the UAV. The UAV can then leverage these solutions to learn the world model and implicitly discover the optimizer's policy in a self-supervised manner. 

\section{Proposed Method}
The proposed method involves two primary steps. Firstly, a world model is learned from demonstrations provided by the TSPWP optimizer. Secondly, the learned world model is used as an internal generative model enriched with active states to simulate the environment and plan actions that minimize the agent's surprise using active inference.

\subsection{World Model}
\begin{figure}[b!]
    \centering
    \includegraphics[width=7.0cm]{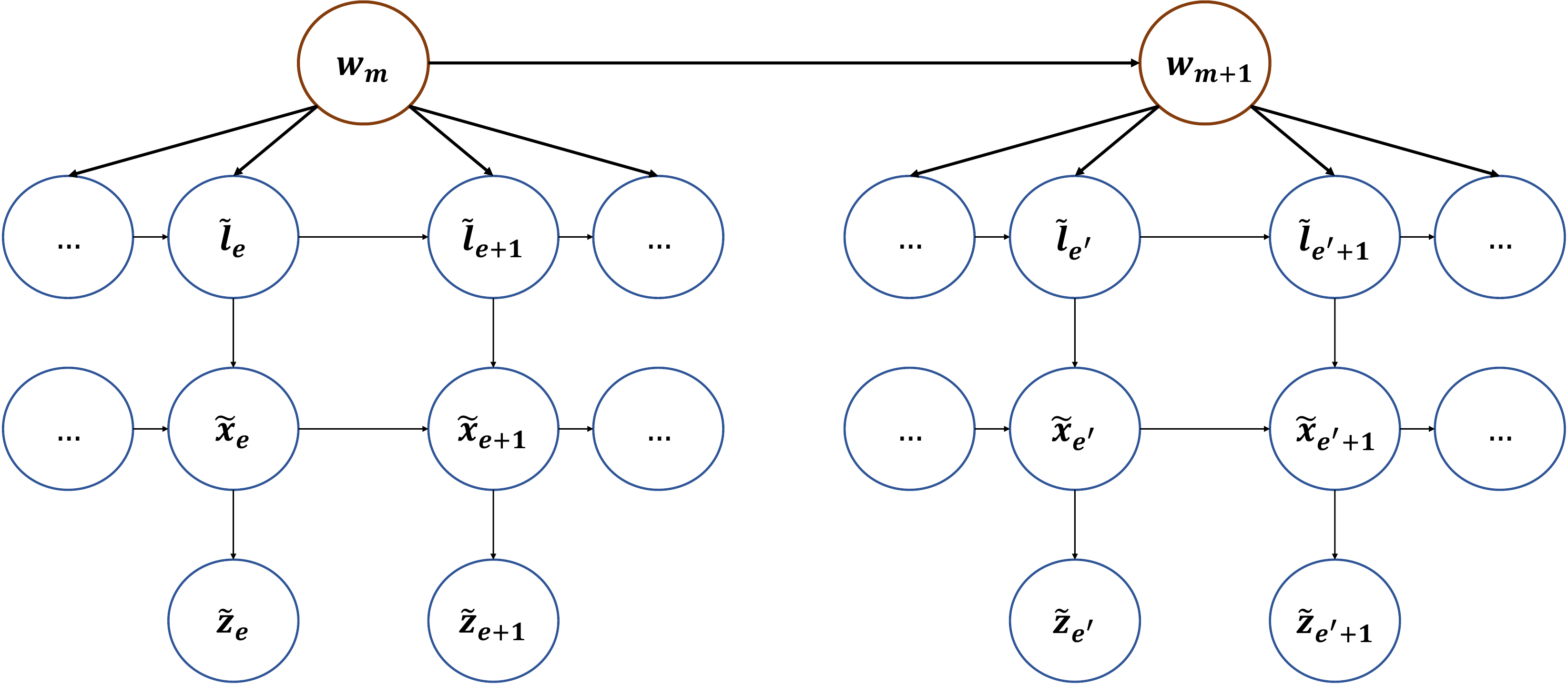}
    \caption{World model structured in a M-GDBN.}
    \label{fig:worldModel_GDBN}
\end{figure}
The wireless network can be visualized as a graph $\mathcal{G}_{i}=(\mathcal{V}_{i}, \mathcal{E}_{i})$ with vertices representing hotspot areas and edges describing possible paths between them. Each vertex $v_{j} =(\mathrm{\boldsymbol{p}}_{j}, R_{j})$, $v_{j} \in \mathcal{V}_{i}$ is associated with a center $\mathrm{\boldsymbol{p}}_{i}$ and an average data rate $R_{i}$, and each edge is associated with a cost $c_{ij}$.
Let $\mathcal{D}=\{\mathcal{G}_{m}\}, m = 1, \dots, M$ be the training set consisting $M$ different realizations (examples) randomly generated. The TSPWP optimizer employs the 2-Opt algorithm \cite{Englert2007WorstCA} to solve the objective function defined in \eqref{TSPWP_objFunc_constraint1}. When given the training set $\mathcal{D}$ as input, the TSPWP optimizer produces a set (representing the output solutions) $\mathcal{L}=\{L_{m}\}$ that encodes the optimal designated sequence of hotspots to solve the $m$-th examples.

Following this, a meta-clustering process was carried out to create a dictionary comprising of words and generalized letters. 
In this dictionary, each individual hotspot is treated as a letter and the tansition between two adjacent letters is regarded as a generalized letter $\Tilde{l}_{i}=[l_{i}, e(l_{i},l_{j})]$ consisting the starting letter $l_{i}$ (i.e., the vertex) and the corresponding derivative $\dot{l}_{i}=e(l_{i}, l_{j})$ (i.e., the outgoing edge ($l_{i}$, $l_{j}$)). Consequently, each word $w_{m}=[\Tilde{l}{i}]$ consists of a sequence of generalized letters describing the visited hotspots in each event. Each word can be seen as a directed graph from which the adjacency matrix can be computed as $\mathbf{A}_{m}=[\mathrm{a}_{ij}]$, where $\mathrm{a}_{ij}=1$ if $(i,j) \in w_{m}$ and $0$ otherwise. In addition, for each formed word ($w_{m}$), we can construct a degree matrix $\mathbf{D}_{m}$ defined as: $\mathbf{D}_{m} = [D_{ij}]$, where $D_{ij}= \sum_{j=1}^{|w_{m}|} \mathrm{a}_{ij}$ if $i=j$ and $0$ otherwise. Accordingly, the transition matrix can be calculated as: $\mathbf{\Pi}_{m}=\mathbf{D}_{m}^{-1}\times\mathbf{A}_{m}$. The global transition matrix ($\mathbf{\Pi}$) that encodes all the letters encountered during training can be estimated by concatenating all the individual matrices ($\mathbf{\Pi}_{1}, \dots, \mathbf{\Pi}_{M}$). The acquired dictionary can be structured in a multi-scale Generalized dynamic Bayesian Network (M-GDBN) as depicted in Fig.~\ref{fig:worldModel_GDBN}. In order to comprehend the generative process forming the optimizers' solutions, we can refer to the dynamic models below:
\begin{subequations}
    \begin{align}
        & w_{m} = \mathrm{f}^{(1)}(w_{m-1}^{}) + \eta_{m}, \label{eq_GDBN1_dynamic_models_level1} \\
        & \Tilde{l}_{e+1} = \mathrm{f}^{(2)}(\Tilde{l}_{e},w_{m}) + \eta_{e+1}, \label{eq_GDBN1_dynamic_models_level2} \\
        & \Tilde{x}_{e+1}^{} = \mathrm{g}^{(1)}(\Tilde{x}_{e}^{l},\Tilde{l}_{e+1}) + \eta_{e+1}, \label{eq_GDBN1_dynamic_models_level3} \\
        & \Tilde{z}_{e+1}^{} = \mathrm{g}^{(2)}(\Tilde{x}_{e+1}^{}) + \nu_{e}. \label{eq_GDBN1_dynamic_models_level4}
    \end{align}
\end{subequations}

\subsection{Decision-making based on active inference}
 
\begin{figure}
    \centering
    \includegraphics[width=4.0cm]{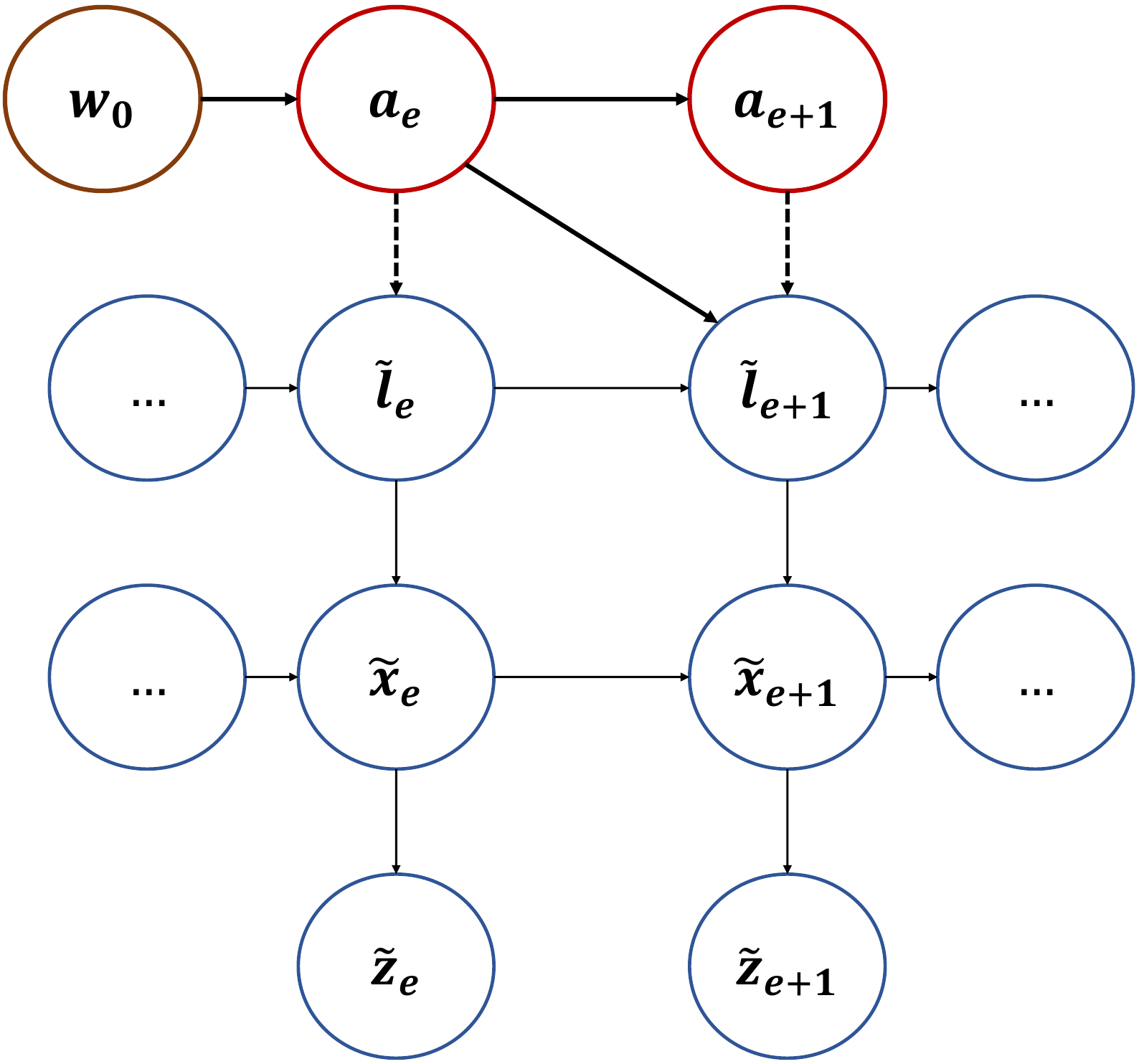}
    \caption{Active M-GDBN.}
    \label{fig:active_GDBN}
\end{figure}
Active inference involves enhancing the world model with active states, creating an active-MGDBN model, shown in Fig.~\ref{fig:active_GDBN}. This model allows the agent to infer hidden environmental states, predict the effects of actions, and anticipate future observations. The active-MGDBN model can be described by a joint distribution function expressed as follows:
\begin{equation}
\scriptsize
    \begin{split}
        \mathrm{Pr}(\Tilde{z}, \Tilde{x}, \Tilde{l}, a, w) = \mathrm{Pr}(w_{m}) \prod_{e=1}^{E} \mathrm{Pr}(\Tilde{z}_{e+1}|\Tilde{x}_{e+1}) \mathrm{Pr}(\Tilde{x}_{e+1}|\Tilde{l}_{e+1}) \\ \mathrm{Pr}(\Tilde{l}_{e+1}|\Tilde{l}_{e}, a_{e}) \mathrm{Pr}(a_{e+1}|a_{e}, w_{m}).
    \end{split}
\end{equation}

During an online mission, a UAV may encounter new scenarios and letters that were not included in its training dataset. To solve a new set of letters represented by $\mathcal{V}_{testing}$ effectively, the UAV needs to connect the vertices optimally, forming the best graph, and find the best possible outcome. Before starting the mission, the UAV plans its actions and relies on the transition matrix to generate probable words that can be used as a reference to complete the online mission successfully. To achieve this, the UAV categorizes the current letters into normal letters (seen during training) and novel letters (not seen during training). Based on the transition matrix, the UAV generates $n$ probable words consisting of only the normal letters as follows:
\begin{equation}
    \mathcal{W}_{testing} = [w^{(1)}, w^{(2)}, \dots, w^{(n)}].
\end{equation}
After generating each word, the UAV determines how similar it is to the words learned and stored in the dictionary (world model). This similarity is calculated using the Levenshtein Distance ($D_{lev}$) \cite{Levenshtein1965BinaryCC}. The generated word with the minimum $D_{lev}$ (i.e., the highest similarity ratio) is selected as the winner:
\begin{equation}
\scriptsize
    w^{(*)} = \min \bigg\{ \min \big\{D_{lev}\big(w^{(1)}, \mathcal{W}\big)\big\}, \dots, \min \big\{D_{lev}\big(w^{(n)}, \mathcal{W}\big)\big\} \bigg\}.
\end{equation}

Upon identifying the most similar generated word, the UAV utilizes it to form an initial (reference) graph $\mathcal{G}_{0}$. Its objective is to expand the reference graph by seamlessly adding new letters. The process involves removing an edge between two existing letters in the reference graph and introducing a new letter, which creates two new edges in the graph. At this stage, the UAV calculates the expected surprise between the new and reference graphs to ensure that the action aligns with the agent's preference. Let's assume there are $p$ letters in the initial graph, and $k$ novel letters are required to cover the current realization. In this scenario, the UAV must add all the new letters while determining the correct order to do so. It adds one letter at a time, considering all possible combinations equivalent to the number of edges ($|\mathcal{E}_{0}|=p$) in the reference graph. Following the order encoded in the new graphs, the UAV can predict the expected sum rate and completion time and compare these predictions with those associated with the reference graph. UAV expresses its belief of how the $p$ planned actions will affect the evolution of the environmental hidden states (representing the sum rate and completion time) at the lower levels using the dynamic model defined in \eqref{eq_GDBN1_dynamic_models_level3} which can be expressed as $\mathrm{Pr}(\Tilde{x}_{e+1}^{(i)}|\Tilde{x}_{e}^{(i)}, \tilde{l}_{e+1}^{(i)}, a_{e}^{(i)})$, $i \in {1,\dots, p}$ and by employing Kalman filter. The posterior representing the predicted state is defined as $\pi(\Tilde{x}_{e+1}^{(i)})=\mathrm{Pr}(\Tilde{x}_{e+1}^{(i)}, \Tilde{l}_{e+1}^{(i)}, a_{e}^{(i)}|\Tilde{z}_{e}^{(i)})$ from which we can obtain the expected observation $\Tilde{z}_{e+1}^{(i)} \sim \mathrm{Pr}(\tilde{z}_{e+1}^{(i)}|\Tilde{x}_{e+1}^{(i)})$ according to the model defined in \eqref{eq_GDBN1_dynamic_models_level4}. Likewise, we can predict the evolution of letters ($\tilde{l}_{e+1}^{(0)}$) and states ($\Tilde{x}_{e+1}^{(0)}$) but which is conditioned on the reference word ($w_{0}$).
The $p$ expected observations (representing the sum-rate and completion time) are compared to the predicted states based on the reference word resulting in $p$ expected surprise indicators calculated using Bhattacharyya distance as follows:
\begin{equation}
    \Upsilon_{i} = - \ln \int \sqrt{\Tilde{x}_{e+1}^{(0)}\times\Tilde{z}_{e+1}^{(i)}} d\Tilde{x}_{e+1}^{(0)}.
\end{equation}
The UAV assesses how surprising it would be to insert a single letter and create a new word. It then picks the word that comes closest to the reference graph regarding completion time and sum rate, resulting in the lowest expected surprise. Therefore, the winning word that causes the most minor expected surprise can be considered an effective physical action, which can be obtained according to:
\begin{equation}
    i = \min \{ \Upsilon_{1}, \Upsilon_{2}, \dots, \Upsilon_{p} \}.
\end{equation}
The UAV will iterate the process until it adds all the novel letters to start the mission. The planning process can also be done online if new letters appear during the mission. The UAV can use the winning graph from the planning process to incorporate the emergent letters and expand it again.

\section{RESULTS AND DISCUSSIONS}
\label{sec:results}
%
%
This section comprehensively evaluates the proposed method's effectiveness in planning a UAV's path to achieve maximum total sum-rate and minimum completion time within a given cell in a purely online manner. Our simulations were conducted in a scenario where a single UAV serves multiple users located in various hotspots across a $2000\times2000$ square meter geographic area.
Simulation paramters are the following: $P_{u}=1$W, $B_{RB}=180$KHz, $\mu_{LoS}=3$, $\mu_{NLoS}=23$, $\sigma^{2}=-104$dBm, $\alpha=0.9$, $\beta=0.1$.
The UAV's altitude was maintained consistently at $200$ meters. During the training phase, we randomly placed $50$ hotspots across the area and used the Poisson distribution to generate user presence and requests within each hotspot. We created a training set, $\mathcal{D}$, comprising M examples of various realizations. Each realization ($m$) consisted of five randomly selected hotspots from the $N_{training}=50$ available hotspots. We utilized the TSPWP optimizer to solve the $M=5000$ examples in $\mathcal{D}$, generating $M$ trajectories and sequences of hotspot visitation order from which we formed a dictionary consisting of a set of words to learn the world model. For our testing phase, we made a total of $N_{testing}=100$ hotspots available, and in each testing example, we randomly chose a certain number of hotspots to be solved by the UAV through pure online means. The UAV relied on the world model that was acquired during training to plan its actions and solve the testing examples. We also evaluated the performance of our proposed method (AIn) against a modified version of conventional Q-learning \cite{articleQL} (modified-QL) that followed the same logic as our proposed method to ensure a fair comparison. To do this, we trained the QL using the same examples that we used to learn the world model where the TSPWP optimizer provided the rewards. In each testing example, we provided QL with the reference word (or graph) to construct a new word based on the current situation using a probabilistic Q-table.

\begin{figure}[htb]
\begin{minipage}[b]{0.48\linewidth}
  \centering
  \centerline{\includegraphics[width=4.0cm]{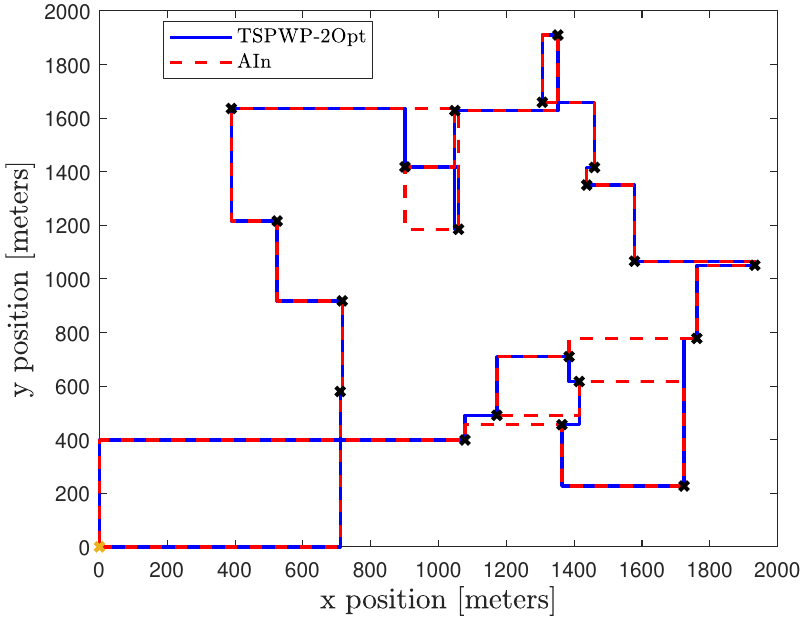}}
  \centerline{(a) AIn (20 Hotspots)}\medskip
\end{minipage}
\begin{minipage}[b]{.48\linewidth}
  \centering
  \centerline{\includegraphics[width=4.0cm]{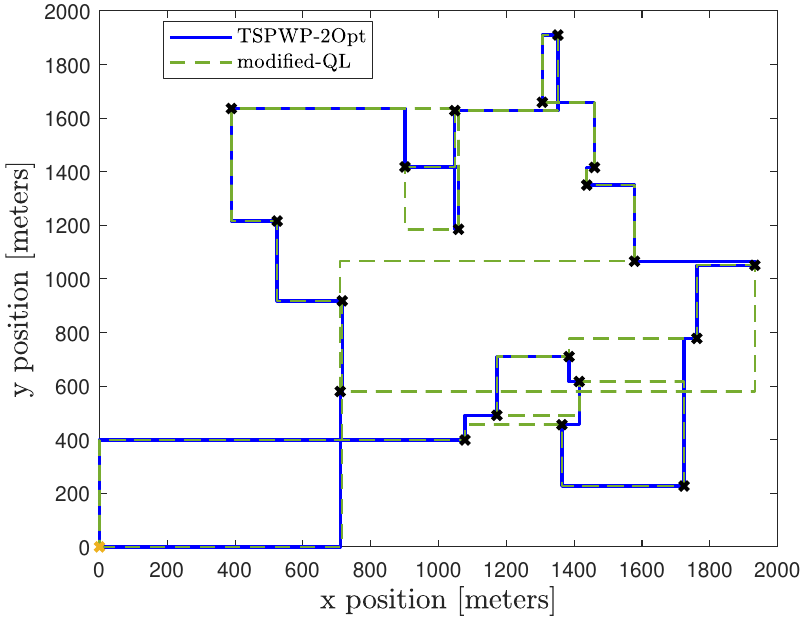}}
  \centerline{(b) modified-QL (20 Hotspots)}\medskip
\end{minipage}
\hfill
\begin{minipage}[b]{0.48\linewidth}
  \centering
  \centerline{\includegraphics[width=4.0cm]{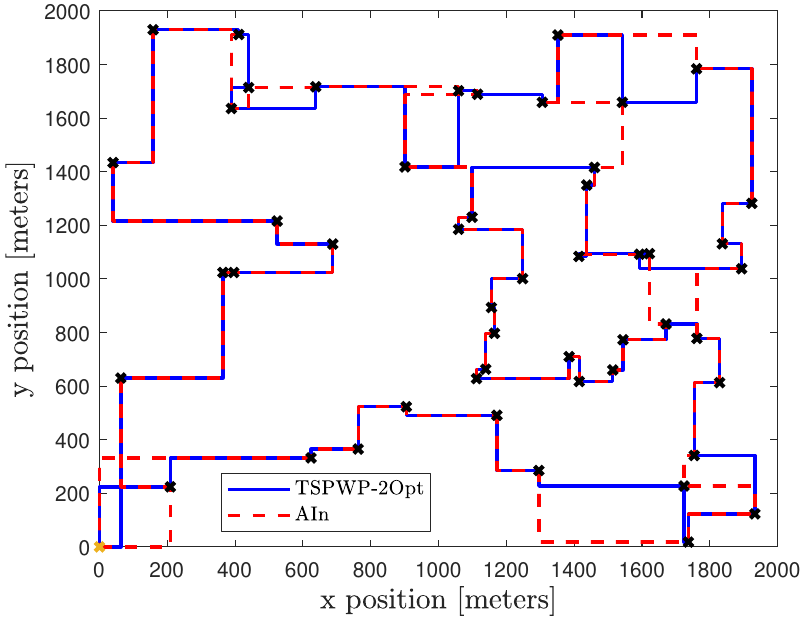}}
  \centerline{(c) AIn (50 Hotspots)}\medskip
\end{minipage}
\hfill
\begin{minipage}[b]{0.48\linewidth}
  \centering
  \centerline{\includegraphics[width=4.0cm]{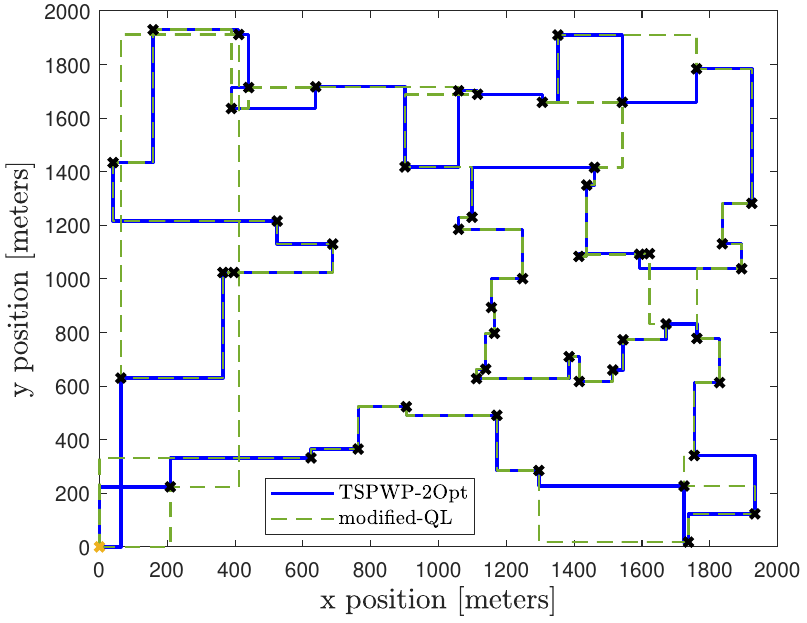}}
  \centerline{(d) modified-QL (50 Hotspots)}\medskip
\end{minipage}
%
\caption{Comparing TSPWP optimizer, AIn, and modified-QL in path planning to solve several testing examples.}
\label{fig:res_trajectories}
\end{figure}

Fig.~\ref{fig:res_trajectories} shows several testing scenarios with different numbers of hotspot areas, along with the trajectories generated by the proposed method (AIn), modified-QL, and the TSPWP optimizer. The AIn approach produces solutions comparable to those of the TSPWP optimizer. This indicates that AIn has successfully captured the optimizer's strategy in a self-supervised manner and generates shorter paths when compared to modified QL.
\begin{figure}[htb]
\begin{minipage}[b]{0.48\linewidth}
  \centering
  \centerline{\includegraphics[width=4.0cm]{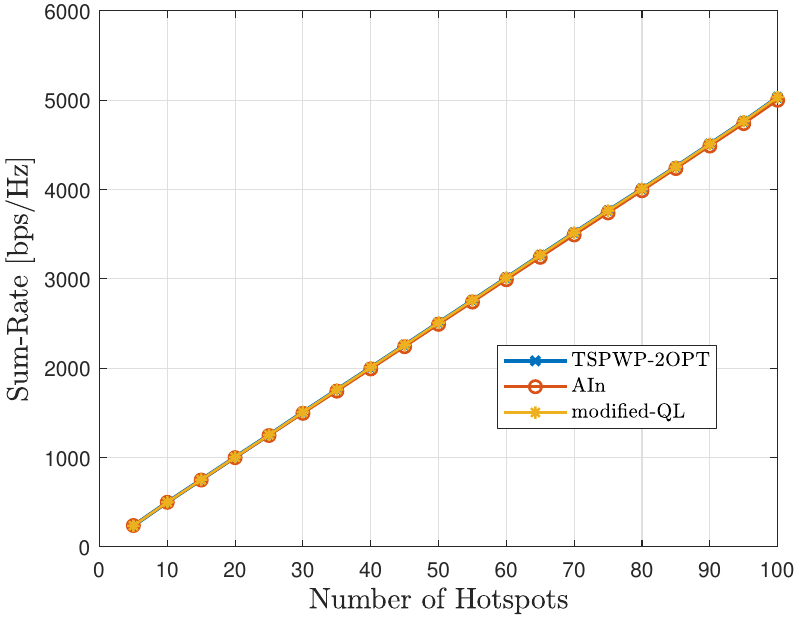}}
  \centerline{(a) Sum-rate}\medskip
\end{minipage}
\hfill
\begin{minipage}[b]{.48\linewidth}
  \centering
  \centerline{\includegraphics[width=4.0cm]{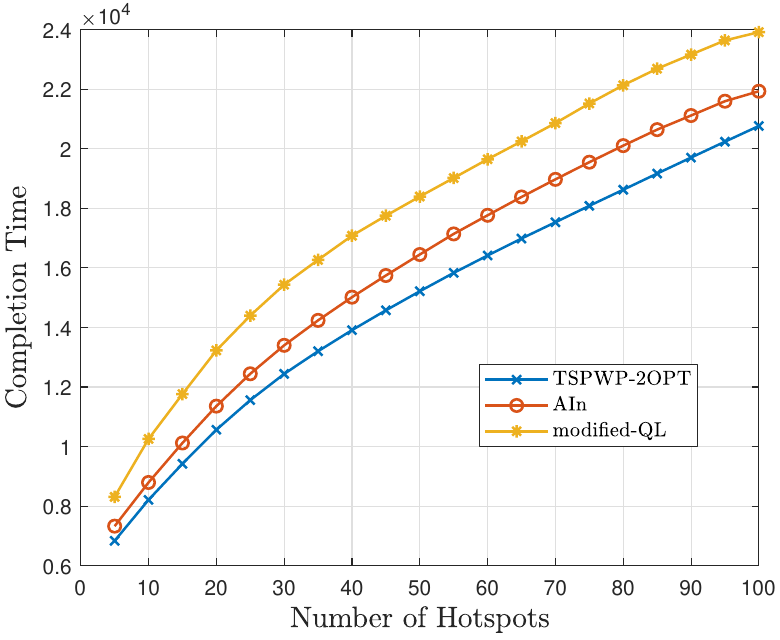}}
  \centerline{(b) Completion time}\medskip
\end{minipage}
\hfill
\caption{Comparing TSPWP optimizer, AIn, and modified-QL performance by varying the number of hotspots.}
\label{fig:sumRate_completionTime}
\end{figure}
\begin{figure}
    \centering
    \includegraphics[width=4.2cm]{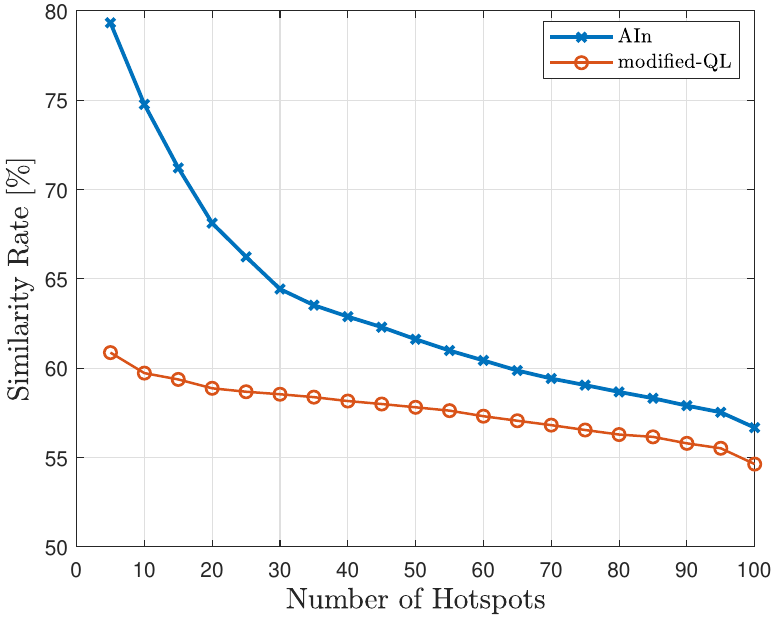}
    \caption{Similarity rate by comparing the words produced by AIn, modified-QL compared to those produced by the TSPWP optimizer.}
    \label{fig:similarityRate}
\end{figure}

In Fig.~\ref{fig:sumRate_completionTime}, we compare the performance of the proposed method with modified-QL in terms of the average sum-rate collected while solving the testing examples. Specifically, we compare it to the analytical sum-rate provided by the TSPWP optimizer. Both AIn and modified-QL approach the optimizer's analytical sum-rate, indicating that both methods successfully visited all the available hotspots during the testing missions. However, it is essential to visit those hotspots in the correct order to minimize the mission's completion time.
In Fig.~\ref{fig:similarityRate}, the completion time of the missions using AIn and modified-QL are compared to the TSPWP optimizer. The proposed approach outperforms the modified QL because it is more flexible in generating feasible solutions that mimic the strategy employed by the optimizer. It is worth noting that the aim is not to replicate the optimizer's strategy precisely by generating the same words (i.e., trajectories) but to comprehend its policy for optimizing the objective function, which was not known to the agent but was implicitly encoded in its internal world model.
Learning from the optimizer's solutions can help the agent understand the underlying rules it follows to solve a particular task. This understanding may produce words that are similar to those produced by the optimizer but not necessarily identical. In Fig.~\ref{fig:similarityRate}, we have compared the similarity ratios between the words generated by the AIn and those produced by the TSPWP optimizer with the similarity between the words produced by the modified-QL and the optimizer's words. The results show that the proposed approach produces words that are more similar to those produced by the optimizer than those generated by the modified-QL. This indicates that the proposed approach follows the optimizer's strategy to some extent and has indirectly understood the objective function. By not producing the same words, the proposed approach can solve problems with greater creativity and enable multiple agents to collaborate and achieve the global goal differently. For instance, this is particularly useful when dealing with a swarm of UAVs.

\section{Conclusion}
This paper proposed a self-supervised method for path planning in UAV-aided networks. It involves learning a world model from expert demonstrations and employing active inference to enable the UAV to make real-time autonomous decisions and engage in online planning. This means considering future observations, scoring policies based on expected surprises, and anticipating outcomes to empower the UAV to take more effective actions. Simulation results have shown that this method provided quicker adaptation to new situations and better performance than traditional Q-learning, leading to broader generalizability.


\vfill\pagebreak

\section{acknowledgement}
This research was partially supported by the European Union’s Horizon Europe research and innovation programme under the Grant Agreement No. 101121134 and by the Italian Ministry of Universities and Research under the National Recovery and Resilience Plan (PNRR) Raise Innovation Ecosystem D.D. 1053 23/6/2022.

\bibliographystyle{IEEEbib}
\bibliography{refs}

\begin{thebibliography}{10}

\bibitem{8579209}
Bin Li, Zesong Fei, and Yan Zhang,
\newblock ``{UAV Communications for 5G and Beyond: Recent Advances and Future Trends},''
\newblock {\em IEEE Internet of Things Journal}, vol. 6, no. 2, pp. 2241--2263, April 2019.

\bibitem{8501974}
Qian Wang, Zhi Chen, Hang Li, and Shaoqian Li,
\newblock ``{Joint Power and Trajectory Design for Physical-Layer Secrecy in the UAV-Aided Mobile Relaying System},''
\newblock {\em IEEE Access}, vol. 6, pp. 62849--62855, 2018.

\bibitem{8531711}
Shuowen Zhang, Yong Zeng, and Rui Zhang,
\newblock ``{Cellular-Enabled UAV Communication: A Connectivity-Constrained Trajectory Optimization Perspective},''
\newblock {\em IEEE Transactions on Communications}, vol. 67, no. 3, pp. 2580--2604, March 2019.

\bibitem{10118877}
Meriem Hammami, Cirine Chaieb, Wessam Ajib, Halima Elbiaze, and Roch Glitho,
\newblock ``{UAV-Assisted Wireless Networks for Stringent Applications: Resource Allocation and Positioning},''
\newblock in {\em 2023 IEEE Wireless Communications and Networking Conference (WCNC)}, March 2023, pp. 1--6.

\bibitem{9234110}
Xiaopeng Yuan, Tianyu Yang, Yulin Hu, Jie Xu, and Anke Schmeink,
\newblock ``{Trajectory Design for UAV-Enabled Multiuser Wireless Power Transfer With Nonlinear Energy Harvesting},''
\newblock {\em IEEE Transactions on Wireless Communications}, vol. 20, no. 2, pp. 1105--1121, Feb 2021.

\bibitem{10086052}
Yunhui Qin, Zhongshan Zhang, Xulong Li, Wei Huangfu, and Haijun Zhang,
\newblock ``{Deep Reinforcement Learning Based Resource Allocation and Trajectory Planning in Integrated Sensing and Communications UAV Network},''
\newblock {\em IEEE Transactions on Wireless Communications}, pp. 1--1, 2023.

\bibitem{9741304}
Ali Krayani, Atm~S. Alam, Lucio Marcenaro, Arumugam Nallanathan, and Carlo Regazzoni,
\newblock ``{An Emergent Self-Awareness Module for Physical Layer Security in Cognitive UAV Radios},''
\newblock {\em IEEE Transactions on Cognitive Communications and Networking}, vol. 8, no. 2, pp. 888--906, June 2022.

\bibitem{9120668}
Xuan Li, Qiang Wang, Jie Liu, and Wenqi Zhang,
\newblock ``{Trajectory Design and Generalization for UAV Enabled Networks:A Deep Reinforcement Learning Approach},''
\newblock in {\em 2020 IEEE Wireless Communications and Networking Conference (WCNC)}, May 2020, pp. 1--6.

\bibitem{10.1162/NECO_a_00912}
Karl Friston, Thomas FitzGerald, Francesco Rigoli, Philipp Schwartenbeck, and Giovanni Pezzulo,
\newblock ``{{Active Inference: A Process Theory}},''
\newblock {\em Neural Computation}, vol. 29, no. 1, pp. 1--49, 01 2017.

\bibitem{friston_2013}
Karl Friston,
\newblock ``{Active inference and free energy},''
\newblock {\em Behavioral and Brain Sciences}, vol. 36, no. 3, pp. 212–213, 2013.

\bibitem{s23156873}
Ali Krayani, Khalid Khan, Lucio Marcenaro, Mario Marchese, and Carlo Regazzoni,
\newblock ``{A Goal-Directed Trajectory Planning Using Active Inference in UAV-Assisted Wireless Networks},''
\newblock {\em Sensors}, vol. 23, no. 15, 2023.

\bibitem{8053918}
Mohammad Mozaffari, Walid Saad, Mehdi Bennis, and Mérouane Debbah,
\newblock ``{Wireless Communication Using Unmanned Aerial Vehicles (UAVs): Optimal Transport Theory for Hover Time Optimization},''
\newblock {\em IEEE Transactions on Wireless Communications}, vol. 16, no. 12, pp. 8052--8066, Dec 2017.

\bibitem{doi:10.1287/trsc.1030.0079}
Dominique Feillet, Pierre Dejax, and Michel Gendreau,
\newblock ``{Traveling Salesman Problems with Profits},''
\newblock {\em Transportation Science}, vol. 39, no. 2, pp. 188--205, 2005.

\bibitem{Englert2007WorstCA}
Matthias Englert, Heiko R{\"o}glin, and Berthold V{\"o}cking,
\newblock ``{Worst Case and Probabilistic Analysis of the 2-Opt Algorithm for the TSP},''
\newblock {\em Algorithmica}, vol. 68, pp. 190--264, 2007.

\bibitem{Levenshtein1965BinaryCC}
Vladimir~I. Levenshtein,
\newblock ``{Binary codes capable of correcting deletions, insertions, and reversals},''
\newblock {\em Soviet physics. Doklady}, vol. 10, pp. 707--710, 1965.

\bibitem{articleQL}
Christopher Watkins and Peter Dayan,
\newblock ``{Technical Note: Q-Learning},''
\newblock {\em Machine Learning}, vol. 8, pp. 279--292, 05 1992.

\end{thebibliography}

\end{document}